\documentclass{article}

\usepackage{arxiv}

\usepackage[utf8]{inputenc} 
\usepackage[T1]{fontenc}    
\usepackage{hyperref}       
\usepackage{url}            
\usepackage{booktabs}       
\usepackage{amsfonts}       
\usepackage{nicefrac}       
\usepackage{microtype}      
\usepackage{amsmath,bm,wrapfig}
\usepackage{graphicx}

\usepackage{xcolor}

\newcommand{\x}{\mathbf{x}}
\newcommand{\y}{\mathbf{y}}
\newcommand{\z}{\mathbf{z}}
\newcommand{\X}{$X$}
\newcommand{\Y}{$Y$}
\newcommand{\Z}{$Z$}
\newcommand{\KL}{\textrm{KL}}
\newcommand{\src}[1]{\textcolor{blue}{(#1)}}

\title{To Beta or Not To Beta: Information Bottleneck for Digital Image Forensics}

\author{%
  Aurobrata Ghosh$^1$ \qquad Zheng Zhong$^1$ \qquad Steve Cruz$^2$ \qquad Subbu Veeravasarapu$^1$ \\
  \textbf{Terrance E Boult}$^2$ \qquad \textbf{Maneesh Singh}$^1$ \\
  $^1$Verisk AI, Verisk Analytics \qquad
  $^2$Vision and Security Technology (VAST) Lab\\
  \texttt{\{aurobrata.ghosh, zheng.zhong, sveeravarasapu, maneesh.singh\}@verisk.com} \\
  \texttt{\{scruz, tboult\}@vast.uccs.edu} \\
}

\begin{document}

\maketitle

\begin{abstract}
	We consider an information theoretic approach to address the problem
	of identifying fake digital images. We propose an innovative method to formulate 
	the issue of localizing manipulated regions in an image 
	as a deep representation learning problem using the Information Bottleneck (IB), 
	which has recently gained popularity as a framework for interpreting deep neural networks. 
	Tampered images pose a serious predicament since digitized media is a 
	ubiquitous part of our lives. 
	These are facilitated by the easy availability of image editing software and aggravated by recent
	advances in deep generative models such as GANs.
	We propose InfoPrint, a computationally efficient solution to the IB formulation using approximate
	variational inference and compare it to a numerical solution that is 
	computationally expensive.
	Testing on a number of standard datasets, we demonstrate that InfoPrint
	outperforms the state-of-the-art and the numerical solution. Additionally, it also 
	has the ability to detect alterations made by inpainting GANs.
\end{abstract}

\section{Introduction}

To a society that is increasingly reliant on digital images and videos as dependable sources 
of information, the ability to photo-realistically alter their contents is a grave danger.
Digital image forensics focuses on this issue by addressing critical problems such as 
establishing the veracity of digital images (manipulation detection), 
pinpointing the tampered regions (manipulation localization), identifying the alteration type, etc.
Different types of alterations call for different forensic techniques. 
Here, we address an important class
of operations that introduce foreign material into an image, e.g., splicing, where part(s)
of other image(s) are inserted, or inpainting, where part(s) are hallucinated by specialized algorithms.
Interestingly, unlike in most computer-vision problems, 
semantic information has had limited success in solving such forensic tasks 
since skilled attackers use semantic structures to hide their modifications.
Instead, non-semantic pixel-level statistics have proven more successful since these
amplify low-level camera-model specific distortions and noise patterns \cite{chen:08}. 
Such digital fingerprints help resolve the integrity of an image by determining 
whether the camera fingerprint is consistent across the entire image.
Over the last decade, a large number of hand engineered low-level statistics have been explored
\cite{zampoglou:16,barni:17,popescu:05}.
However, with the technological improvement of image editing/altering software and recent
deep generative models, forensic algorithms need to make commensurate advances
towards data-driven deep learning solutions.

In this work, we propose to use the information bottleneck (IB) \cite{tishby:99}
to cast the problem of modelling the distinguishing low-level camera-model statistics (fingerprint) 
as a data-driven representation learning problem. 
Working with image patches, we design an encoder-decoder architecture, where we control
the information flow between the input-patches and the representation layer. 
This constriction of the mutual information allows our network to ignore the unimportant semantic 
information contained in any patch and focus its capacity to learning only the useful 
features needed to classify the source camera-model. 
This application of the IB principle is, in a certain sense, 
the complete reverse of  why it has been typically applied \cite{alemi:17,alemi:18}. 
We use it to learn useful residual "noise" patterns and ignore the semantics rather than 
the other way around. Since the learned noise pattern representation is like a 
camera-model's fingerprint, we call our proposed method InfoPrint (IP).

The main contributions of this work are the following.
First, we propose the IB formulation, which converts a classical
feature modelling problem for identifying camera-models into a deep representation learning problem.
To the best of our knowledge, this is a unique application of IB to a growing 
real-world problem with serious consequences.
Our IB application is also novel in the sense that it encourages learning low-level noise patterns 
rather than semantic information, which is the opposite of how IB is normally used.
Second, we propose a computationally efficient approximate solution based on variational IB 
\cite{alemi:17} and compare it to a standard and expensive numerical estimation to show that 
the variational solution can outperform the numerical one. This shows that this representation learning
problem can be solved numerically or approximated using variational inference and that the latter
is sufficiently good to outperform the former at the task of splice localization.
It also allows us to effectively explore relatively deep models and long training procedures.
Third, by conducting experiments on a suite of standard test datasets, we demonstrate that
our method outperforms the state-of-the-art by up to $5\%$ points. 
Finally, we show InfoPrint's ability to detect the signatures of deep generative
models by pitting it against three state-of-the-art inpainting GANs.


\section{Related Work}

\noindent
\textbf{Image forensics}\quad
The image formation process broadly consists of three stages: sensor measurements, in-camera
processing and storage, which may include compression. 
This process is unique in every camera-model and leaves subtle distortions
and noise-patterns in the image, which are invisible to the eye. 
Since these traces are specific to every camera-model, 
they have been researched for forensic applications.
Sensor pattern noise, which originates from imperfections in the sensor itself, was explored in
\cite{lukas:06b}. It was shown to be sensitive to all manipulation types and was used for both
the detection and localization of forgeries. However, this pattern noise is difficult to detect 
in regions with high texture and is absent, or suppressed, in saturated and dark regions of an image.
Colour filter array (CFA) demosaicking is an in-camera processing step that produces the pixel
colours. Different detection and localization  strategies based on  CFA signature inconsistencies 
were proposed in \cite{dirik:09,popescu:05}. However, the scope of such specialized CFA models 
is often limited. JPEG is the most common storage format, and it too 
carries signatures of camera models, such as dimples \cite{agarwal:17}, 
or can contain clues about post-processing steps, such as traces of multiple compressions 
\cite{barni:17,zampoglou:16}. Although JPEG statistics have been explored for both detection
and localization tasks, these are format specific and do not generalize to other common, or new, formats.

More generic approaches include modelling noise-residuals, 
which are statistical patterns not attached to a specific source but are the result
of the combined processes of the whole imaging pipeline.
These can be discerned by suppressing the semantic contents of
the image. For example, \cite{mahdian:09} used the wavelet transform as a high-pass filter to estimate
the noise-residuals and then determined its inconsistencies. 
Spatial rich filters (RF) \cite{fridrich:12}, a set of alternate high-pass filters to model the
local noise-residuals, are also employed widely. While \cite{cozzolino:15} looked at the co-occurrences
of one RF, \cite{zhou:18} used three residual filters along with the colour information in a 
convolutional neural network (CNN) to localize forgeries. Learned RFs using constrained convolutions
were employed in \cite{mayer:18} for the localization task. 
Noiseprint \cite{cozzolino:18b}, a novel approach, used a denoising CNN to estimate 
the properties of the noise-residuals and changes in these, to discover forgeries. 
In \cite{bondi:17b}, a CNN was trained for camera-model 
identification to discover forgeries, but it did not exploit noise-residuals and relied on a CNN
architecture effective for learning semantic contents.
Here, we propose to use a constrained convolution layer mimicking RFs and IB to learn noise-residual
patterns and localize manipulations.
An idea related to this work was submitted in a workshop, which we 
include in the supplementary material \cite{ghosh:19}.
It used a mutual information-based regularization that was computed numerically through binning.
We improve upon that idea by showing that IB provides a formal framework to interpret the 
regularization, which permits a more efficient solution using variational approximation. Additionally,
it allows us to tune the regularization in a principled manner, which enhances the forensic performance. 

\noindent
\textbf{Mutual Information \& Information Bottleneck}\quad
Information theory is a powerful framework that is being increasingly adopted to improve 
various aspects of deep machine learning, e.g., representation learning \cite{hjelm:19}
generalizability \& regularization \cite{peng:19}, and for interpreting how deep neural networks 
function \cite{shwartz:17,michael:18}. Mutual information plays a key role in many of these methods.
InfoGAN \cite{chen:16}, showed that maximizing the mutual information 
between the latent code and the generator's output
improved the representations learned by a generative adversarial network (GAN) \cite{goodfellow:2014},
allowing them to be more disentangled and interpretable. Since mutual information is hard to compute,
InfoGAN maximized a variational lower bound \cite{barber:03}.
A similar information maximization idea was explored in \cite{hjelm:19} to improve unsupervised 
representation learning using the numerical estimator proposed in \cite{belghazi:18}.

Information bottleneck \cite{tishby:99} curtails the information flow between 
the input and a representation layer, which encourages a model to learn task related 
features and helps improve its generalization ability.
However, since the IB Lagrangian is hard to solve in practice, variational approximations
suitable for deep learning were proposed in \cite{alemi:17,achille:18}, which also showed
that IB is closely related to variational autoencoders \cite{kingma:13} (VAEs). 
Additionally, \cite{achille:18} proved that IB could be used to learn disentangled representations.
Similar bottleneck ideas to improve the disentanglement of representations learned by VAEs 
were investigated empirically in \cite{higgins:17,burgess:18}.
An insightful rate-distortion interpretation using IB was applied to VAEs in \cite{alemi:18}.
Recently, in \cite{peng:19}, IB was proposed as an effective regularization and shown
to improve imitation learning, reinforcement learning, and the training of GANs.
Here, we leverage the variational IB formulation that was developed for deep neural networks 
in \cite{alemi:17,achille:18} using the reparameterization trick of \cite{kingma:13}.

\section{Preliminaries}

We briefly review the IB framework and its variational approximation.
Learning a predictive model $p(\y|\x)$ is hampered when a model overfits nuisance detractors that exist
in the input data \X, instead of focusing on the relevant information for a task \Y.
This is especially important in deep learning when the input is high dimensional (e.g., an image),
the task is a simple low dimensional class label, and the model is a flexible neural network.
The goal of IB is to overcome this problem by learning a compressed 
representation \Z, of \X, which is optimal for the task \Y~in terms of mutual information.
It is applied by maximizing the IB Lagrangian \cite{tishby:99} 
based on the mutual informationn values $I(Z,X)$, 
$I(Z,Y)$ (we follow the convention in \cite{alemi:17}):
\begin{equation}
\mathcal{L} = I(Z,Y) - \beta I(Z,X).
\label{eq:IB}
\end{equation}
By penalizing the information flow between \X~and \Z~while 
maximizing the mutual information required for the task, 
IB extracts the relevant information that \X~contains about \Y~and discards non-informative 
signals. This leads to learning a representation \Z~with an improved generalization ability.

However, since mutual information is hard to compute in a general setting, especially with high dimensional variables, \cite{alemi:17,achille:18} proposed a variational approximation 
that can be applied to neural networks. Let \Z~be a stochastic encoding layer, then
from the definition of mutual information:
\begin{equation}
I(Z,Y) = \int p(\y,\z)\log\frac{p(\y,\z)}{p(\y)p(\z)} d\y d\z =
\int p(\y,\z)\log p(\y|\z) d\y d\z - \int p(\y)\log p(\y) d\y,
\end{equation}
where the last term is ignored as it is the entropy of $\y$ and is constant. 
In the other term, $p(\y|\z)$ is intractable and is approximated using a 
variational distribution $q(\y|\z)$, the decoder network. 
Then, a lower bound of $I(Z,Y)$ is found because the KL divergence
$\KL[p(\y|\z)||q(\y|\z)]\ge0 \implies \int p(\y|\z)\log p(\y|\z)d\y \ge \int p(\y|\z) \log q(\y|\z) d\y$
and by assuming a Markov chain relation $Y \rightarrow X \rightarrow Z$: 
\begin{equation}
I(Z,Y) \ge \mathbb{E}_{\x,\y\sim p(\x,\y)}\left[ \mathbb{E}_{\z\sim p(\z|\x)}
\left[ \log q(\y|\z) \right]\right],
\label{eq:CE}
\end{equation}
where $p(\z|\x)$ is an encoder network and $p(\x,\y)$ can be approximated using the training data 
distribution. Therefore, the r.h.s. of Eq~\ref{eq:CE} becomes the average cross-entropy (with stochastic
sampling over $\z$). 
Proceeding similarly:
\begin{equation}
I(Z,X) = \int p(\x,\z)\log\frac{p(\x,\z)}{p(\x)p(\z)}d\x d\z =
\int p(\x,\z)\log p(\z|\x) d\x d\z - \int p(\z)\log p(\z)d\z.
\end{equation}
In this case, $p(\z)$ is intractable and is approximated by a prior marginal distribution $r(\z)$.
An upper bound for $I(Z,X)$ is found because
$\KL[p(\z)||r(\z)]\ge0 \implies \int p(\z)\log p(\z)d\z \ge \int p(\z) \log r(\z) d\z$, therefore:
\begin{equation}
I(Z,X) \le \int p(\x)p(\z|\x)\log\frac{p(\z|\x)}{r(\z)} d\x d\z= 
\mathbb{E}_{\x\sim p(\x)}\left[ KL[p(\z|\x)||r(\z)] \right].
\label{eq:IBReg}
\end{equation}
Again, $p(\x)$ can be approximated using the data distribution. 
Replacing Eqs~\ref{eq:CE},\ref{eq:IBReg} in Eq~\ref{eq:IB} gives the variational IB objective function:
\begin{equation}
J_{IB}(p, q) =  \frac{1}{N}\sum_{i=1}^N \mathbb{E}_{\z\sim p(\z|\x_i)}\left[-\log q(\y_i|\z)\right]
+ \beta~\KL[p(\z|\x_i)||r(\z)] \ge -\mathcal{L},
\label{eq:VIB}
\end{equation}
which can be minimized using the reparameterization trick of \cite{kingma:13}.
According to the rate-distortion interpretation of IB \cite{alemi:18}, the loss
term is denoted as distortion $D$ which approximates the non-constant part of $-I(Z,Y)$, 
while the unweighted regularization term is denoted as rate $R$ which approximates $I(Z,X)$.
$R$ measures the excess number of bits required to encode representations. The $RD$-plane allows
to visualize the family of solutions to the IB Lagrangian for different values of $\beta$ and provides
insight into the encoder-decoder network's properties.


\section{Proposed Method}

Our goal is to localize a digital tampering that inserts foreign material into a host image to
alter its contents. Since such splicing operations are often camouflaged by semantic structures, we
assume that we can determine such forgeries by inspecting low-level pixel statistics. In general,
splices will contain different statistical fingerprints than the host image as they are likely to 
originate from either a different camera-model or a different image formation process, e.g., inpainting.
Such an assumption has broad application scope and is widely used to design forensic methods
\cite{cozzolino:15,cozzolino:18b,huh:18}. 

To achieve this goal, we expand prior strategies \cite{bondi:17b,cozzolino:18b}, to use our novel IB-based loss to learn to ignore semantic content. 
First, we train a deep neural network to learn a representation that captures the statistical
fingerprint of the source camera-model from an input image patch while ignoring the semantic content. 
Next, we compute the fingerprint-representation for different parts of a test image. 
Finally, we look for inconsistencies in these to localize splicing forgeries.
An important point is that we train our network with a large number of camera models to improve its ability
to segregate even unseen devices. This allows our network to be effective in a 
blind test setting where we apply it on images acquired on unknown cameras.

\textit{A. Learning representations}

Learning low level representations consists of two steps: extracting noise-residuals and 
learning suitable representations from these. To extract noise-residuals we consider learned local noise
models. In particular, we consider a constrained convolution layer of the form \cite{ghosh:19}:
\begin{equation}
\mathcal{R}^{(k)} = \mathbf{w}_k(0,0) + \sum_{i,j\neq0,0}\mathbf{w}_k(i,j) = 0,
\label{eq:RF}
\end{equation}
which binds the weights of the $k^{\textrm{th}}$ filter to compute the mismatch, or noise-residual,
between a pixel's value at position $(0,0)$ and its value as interpolated from its 
$S\times S$ neighbours.
These are high-pass filters similar to RFs \cite{fridrich:12} that 
model noise-residuals locally by suppressing the semantic contents and can be trained end-to-end
by including the penalty $\mathcal{R}=(\sum_{k}(\mathcal{R}^{(k)})^2)^{\frac{1}{2}}$ in the optimization.

However, since these noise models are high-pass filters, they also capture high-frequency semantic contents, such as edges and textures, which carry scene-related information we seek to ignore.
The ideal noise-residuals should be high-frequency content uncorrelated to the semantic information.
To learn these, one could regularize the mutual information between the input and the feature layer in a neural network \cite{ghosh:19}.
Intuitively, that would discourage the correlation between the learned features and semantic contents in the input.
However, mutual information is notoriously hard to compute.
A numerically expensive binning method was used in \cite{ghosh:19} that curtailed the training process.
Here, we re-interpret the mutual information regularization through the IB framework.
This not only allows us to employ an efficient variational solution and explore longer training processes but also provides us with the $RD$-plane, which we inspect to select the best regularization parameter $\beta$ (Eq~\ref{eq:VIB}).

To learn suitable representations using IB, we adopt a stochastic encoder-decoder architecture. 
For input, we consider \X~to be an image patch, \Y~to be a class label for the
task of camera-model identification and \Z~to be a stochastic encoding layer. 
We select a classification task because it
is aligned with our intent to learn non-semantic features from the input since the semantic contents of an image 
cannot be related to the camera-model in an obvious way.
Additionally, we are able to exploit large camera-tagged untampered image databases for training. 
This allows us to avoid specialized manipulated datasets and avert the chances of overfitting to these. 
With this setting, we are able to train our network by simply minimizing the variational IB objective 
function in Eq~\ref{eq:VIB}.

\begin{table}[t!]
	\caption{InfoPrint architecture. A CNN inspired by ResNet-18v1 \cite{he:16}, with 27 layers.
		All convolutions have stride 1. The input patch size is $49\times49\times3$ and the output encoding size 
		is $1\times1\times72$. }
	\label{table:architecture}
	\centering
	\footnotesize
	\begin{tabular}{ccc}
		\toprule
		constrained conv. & residual conv. & encoding conv. \\ \midrule\midrule
		$3\times3,64$  \qquad &
		$\underbrace{
			7\times7,64,~
			\underbrace{\left[ \begin{matrix}
				3\times3,64 \\
				3\times3,64
				\end{matrix} \right]}_{\textrm{res.block}},~
			5\times5,64, ~
			\underbrace{\left[ \begin{matrix}
				3\times3,64 \\
				3\times3,64
				\end{matrix} \right]}_{\textrm{res.block}}}_{\times4},~
		7\times7,64$\qquad
		&
		$1\times1,(\underbrace{36+36}_{\bm{\mu},~\bm{\sigma}})$ \\
		\bottomrule
	\end{tabular}
\end{table}

For our encoder $p(\z|\x)$, we propose an architecture that is inspired by ResNet-18 version-1 \cite{he:16},
where we include an initial constrained convolution layer (Eq~\ref{eq:RF}) to model noise-residuals and
where we discard operations that quickly shrink the input and encourage learning high level (semantic) features. 
Namely, we discard the initial max-pooling layer, all convolutions with stride greater than one, and the final
global average pooling layer. We found these to be detrimental to our task.
However, since we were keen on ending the network with a single ``feature-pixel'' with a large bank of filters
(to avoid fully connected layers),
we insert additional $7\times7$ and $5\times5$ convolutions. 
The final architecture,  detailed in Table~\ref{table:architecture}, is a 27-layers deep CNN where every 
convolution is followed by batch normalization and ReLU activation.
To get a stochastic encoding \Z, we split the CNN's output vector of 72 filters into $\bm{\mu}_{\x}$ and $\bm{\sigma}_{\x}$ 
and model $p(\z|\x) = \mathcal{N}(\bm{\mu}_{\x},diag(\bm{\sigma}_{\x}))$ \cite{kingma:13}.

We adopt an extremely simple decoder $q(\y|\z)$ to deter our model from degenerating to the auto-decoder limit, 
an issue that is faced by VAEs with powerful decoders \cite{alemi:18}. In fact, we also observed this issue while designing
our decoder. Hence, following \cite{alemi:17}, 
we select a simple logistic regression model: a dense (logit generating) layer
that is connected to the stochastic code layer \Z~and is activated by the softmax function.

To select the regularization parameter $\beta$ in a principled manner, we turn to 
the $RD$-plane \cite{alemi:18} to gain insights into the encoder-decoder's characteristics. 
There exists an $RD$ \textit{curve} that divides the plane into practical feasible and infeasible regions. 
Inspecting this curve allows selecting $\beta$ to balance the trade-off between
the distortion, which affects task accuracy, and the rate, which affects compression 
and hence the generalization capacity. However, we have two tasks.
Although our main task is splice localization since manipulated datasets are limited,
we train our model on a secondary task of camera-model identification. 
Therefore, we employ the $RD$ curve of the training task to first identify the potential range for $\beta$, 
then we select the optimal $\beta$('s) from this range through empirical testing.

\textit{B. Forgery localization}

Assuming that the untampered region is the largest part of the image like in \cite{cozzolino:18b,cozzolino:15},
we simplify the splice localization problem to a two-class feature segmentation problem. First,
we compute our network's representation $(\bm{\mu},\bm{\sigma})$ as a predictive signature of the camera-model for
all juxtaposed patches in the test image. Then, following \cite{cozzolino:18b,cozzolino:15},
we segment these using a Gaussian mixture model with two components using expectation maximization (EM). The Gaussian
distributions are only approximate statistics of the features of the two classes and help to separate them 
probabilistically. Note that forgery detection cannot be performed by our method since it will always identify two
classes. However, that is a separate problem that can be solved using other methods like \cite{barni:17}.

\textit{C. Implementation}

In our experiments, we consider input patches of size $49\times49\times3$ and $k=64$ constrained convolutions
with support $S=3$ in the first layer. Also, our encoder has a fixed number of 64 filters in every layer, unlike in the 
original ResNet. For the variational prior distribution, we use the factorized standard Gaussian
$r(\z)=\prod_i\mathcal{N}_i(0,1)$ proposed in \cite{alemi:17} and train our network using the loss:
\begin{equation}
J = J_{IB} + \lambda\mathcal{R} + \omega_1||\mathbf{W}||_1 + \omega_2||\mathbf{W}||_2,
\end{equation}
where $\mathbf{W}$ are all weights of the network and we empirically select $\lambda=1$ and $\omega_1=\omega_2=1e-4$.

\begin{table}[t!]
	\caption{F1 \& MCC scores on the test datasets. Black: optimal threshold, blue: Otsu threshold}
	\label{table:F1-MCC}
	\centering
	\footnotesize
	\begin{tabular}{lllllllll}
		\toprule
		(F1) &  DSO-1 & NC16 & NC17-dev1 & \qquad& (MCC) &  DSO-1 & NC16 & NC17-dev1 \\  \midrule \midrule
		NoMI & 0.64 \src{0.52}  & 0.39  \src{0.29}& 0.39  \src{0.31} & & NoMI & 0.60 \src{0.47}  & 0.37  \src{0.27}& 0.32  \src{0.22} \\
		MI \cite{ghosh:19} & 0.69 \src{0.59} & 0.39 \src{0.28} & 0.40  \src{0.32} & & MI \cite{ghosh:19} & 0.65 \src{0.55} & 0.37 \src{0.26} & 0.33  \src{0.25}    \\
		\textbf{IP1e-3} & 0.71 \src{0.55}  & \textbf{0.42} \src{0.29} & \textbf{0.44} \src{0.31} & & \textbf{IP1e-3} & 0.67 \src{0.53}  & \textbf{0.40} \src{0.28} & \textbf{0.38} \src{0.25} \\
		\textbf{IP5e-4} & \textbf{0.72} \src{0.58}  & 0.40 \src{0.29} & 0.42 \src{0.31} & & \textbf{IP5e-4} & \textbf{0.69} \src{0.55}  & 0.38 \src{0.27} & 0.35 \src{0.24} \\
		SB & 0.66 \src{0.54} & 0.37 \src{0.26} & 0.43 \src{0.36} & & SB & 0.61 \src{0.48} & 0.34 \src{0.25} & 0.36 \src{0.25} \\
		EX-SC & 0.57 \src{0.49}  & 0.38 \src{0.31} & \textbf{0.44} \src{0.37} & & EX-SC & 0.52 \src{0.43}  & 0.36 \src{0.29} & \textbf{0.38} \src{0.30} \\
		\bottomrule
	\end{tabular}
\end{table}

\begin{wraptable}{l}{7cm}
	\caption{AUC scores on the test datasets. }
	\label{table:AUC}
	\centering
	\begin{tabular}{llll}
		\toprule
		(AUC) &  DSO-1 & NC16 & NC17-dev1 \\  \midrule \midrule
		NoMI & 0.87   & 0.80  & 0.80      \\
		MI \cite{ghosh:19} & 0.90  & 0.81  & 0.80      \\
		\textbf{IP1e-3} & \textbf{0.92}  & \textbf{0.83} & \textbf{0.82}  \\
		\textbf{IP5e-4} & \textbf{0.92}  & \textbf{0.83} & \textbf{0.82}  \\
		SB & 0.86  & 0.77  & 0.77  \\
		EX-SC & 0.85   & 0.80  & 0.81  \\
		\bottomrule
	\end{tabular}
\end{wraptable}


For training, we use the Dresden Image Database \cite{gloe:10} that contains more than 17,000 JPEG images coming
from 27 camera models. For each camera-model, we randomly select $70\%$ of the images for training, $20\%$ for validation, and $10\%$ for testing. We use a mini-batch of 200 patches, and train for 700 epochs with 100,000 randomly selected
patches in every epoch. We maintain a constant learning rate of $1e-4$ for 100 epochs, then decay it linearly to
$5e-6$ in the next 530 epochs and then finally decay it exponentially by a factor 0.9 over the last 70 epochs.
This allows us to achieve a camera-model prediction accuracy of $\sim80\%$ on the validation and test sets for various 
values of $\beta$.

We used TensorFlow for our implementation and trained on NVIDIA Tesla V100-SXM2 (16GB) GPUs with 32 CPUs and 240GB RAM.
To compare, we also trained the 18 layers deep network in \cite{ghosh:19} with 64 filters (instead of 19) and
$72\times72\times3$ input patches for the same number of epochs, although we had to decrease the batch size to 100. 
We trained two models: with (MI) \cite{ghosh:19} and without (NoMI) mutual information regularization.
While training our variational model took about 14 hours, training the MI model took eight days in comparison, indicating
the efficiency of the variational solution in contrast to the numerically expensive binning method (MI) \cite{ghosh:19}.

\section{Experiments \& Results}

To evaluate our method, we stress test it on three standard manipulated datasets and report
scores using three metrics. However, first, we tune our model by running detailed experiments
to explore the $RD$ curve and select the optimal regularization parameter $\beta$.
Then, we conduct an ablation study to gauge the relevance of IB and finally compare InfoPrint to two state-of-the-art algorithms. 

The manipulation datasets we employ are DSO-1 \cite{carvalho:13}, Nimble Challenge 2016 (NC16)
and the Nimble Challenge 2017 (NC17-dev1) \cite{medifor:17}. DSO-1 consists of 100 spliced
images in PNG format, where the tampered regions are relatively large but well camouflaged by the semantic
contents of the image. NC16 contains 564 spliced images mostly in JPEG format.
NC17 contains 1191 images with different types of manipulations. Of these, 
we select only the spliced images, which amount to 237. NC16 and NC17-dev1 images
contain a series of manipulations, some of which are complex operations that attempt to erase traces of forgeries. Furthermore, the tampered regions are often small.
All three datasets are state-of-the-art, contain hard to detect forgeries and are accompanied by 
the ground truth manipulation masks.
Additionally, we generate manipulations created by three inpainting GANs, namely 
Yu et al. \cite{yu:18}, Nazeri et al. \cite{nazeri:19}, Liu et al. \cite{liu:18},
which represent the state-of-the-art in generative models for image inpainting.

To score our performance, we use the F1 score, Matthews Correlation Coefficient (MCC), and area under the receiver operating 
characteristic curve (ROC-AUC). These are widely used metrics for evaluating splice localization 
\cite{zampoglou:16,cozzolino:18b}. However, F1 and MCC require a binarized forgery prediction mask, while 
our method predicts probabilities from the EM segmentation. Although in the forensic literature it is
customary to report the scores for optimal thresholds, computed from the ground truth masks, we
additionally report scores from automatic thresholding using Otsu's method \cite{otsu:79}.

For comparison with InfoPrint, we consider two ablated models: without mutual information regularization (NoMI), 
and with mutual information (MI) computed using the binning approach in \cite{ghosh:19}. 
These were described earlier.
While selecting the optimal $\beta$, we also see a variational model with no IB regularization ($\beta=0$).
These help to gauge the importance of information regularization and compare the expensive numerical solution
of \cite{ghosh:19} to the efficient variational solution proposed here. Additionally, we consider SpliceBuster 
\cite{cozzolino:15}, a state-of-the-art splice localization algorithm that is a top-performer of the NC17 challenge \cite{medifor:17},
and EX-SC \cite{huh:18}, a recent deep learning based algorithm that predicts meta-data self-consistency to 
localize tampered regions.

\begin{figure}
	\centering
	\includegraphics[width=0.9\linewidth]{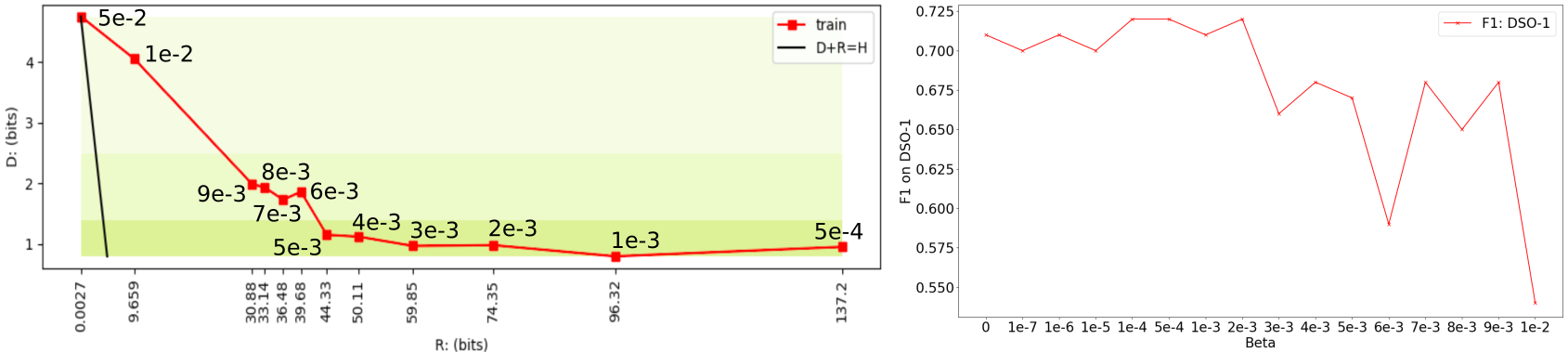}
	\caption{$\beta$ selection from $RD$ curve. Left: $RD$ curve in red, right: F1 metric on DSO-1.
		Low distortion values are attained for $\beta \le 5e-3$, while a peak in the F1 is observed from 2e-3 to 1e-4.}
	\label{fig:RD}
\end{figure}

To select $\beta$ we plot the $RD$ curve (Fig~\ref{fig:RD}). We observe that our model
achieves low distortion values for $\beta \le 5e-3$ for the training task. To select $\beta$ for the forensic
task, we compute F1 scores on DSO-1 for all values of $\beta$ till 0 and find a peak from 2e-3 to 1e-4
(1e-3 is an anomaly we attribute to stochastic training).
Hence we conduct our experiments for two central values, $\beta=1e-3,~5e-4$.

Quantitative results for splice localization are presented in Tables~\ref{table:F1-MCC}\&\ref{table:AUC}
and include results for our proposed model, ablated models, and state-of-the-art models.
All three scores presented in these tables indicate improved results over published methods.
The F1 scores indicate up to $6\%$ points improvement over SB and $15\%$ points improvement over EX-SC
on DSO-1 and best scores on NC16 \& NC17.
The MCC scores are again high for InfoPrint in comparison to the other methods, with a margin of up to $8\%$
points on DSO-1 in comparison to SB. The trend for AUC scores is similar.
Comparing the performances of the ablated models indicates that information regularization greatly
improves forensic performance. Furthermore, it is also clear that the variational solution InfoPrint
improves over the performance of the numerically expensive MI model \cite{ghosh:19}.

\begin{figure}
	\centering
	\includegraphics[width=1\linewidth]{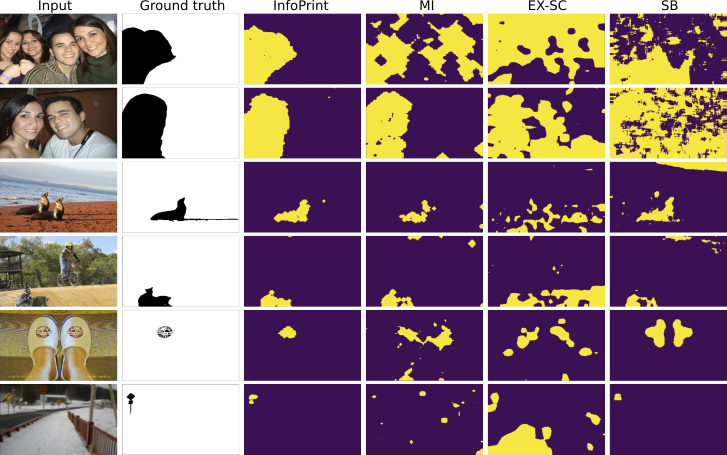}
	\caption{Qualitative results showing the superiority of InfoPrint over published methods.}
	\label{fig:comp}
\end{figure}

\begin{figure}
	\centering
	\includegraphics[width=1\linewidth]{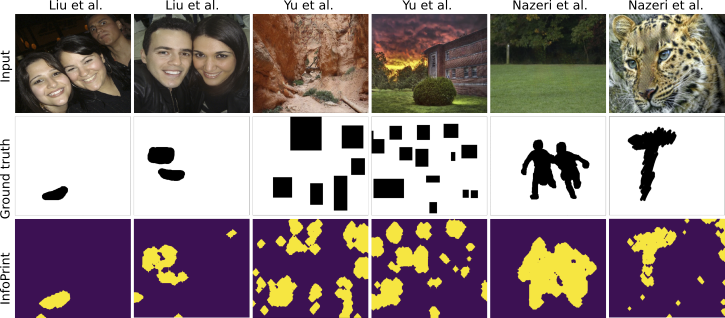}
	\caption{InfoPrint vs inpainting GANs: Liu et al. \cite{liu:18}, Yu et al. \cite{yu:18}, Nazeri et al. \cite{nazeri:19}.}
	\label{fig:GANs}
\end{figure}

Qualitative results are presented in Figs~\ref{fig:comp}\&\ref{fig:GANs}. Fig~\ref{fig:comp} compares
InfoPrint's predicted manipulation mask to the ground truth mask and masks predicted by MI and the 
published methods SB and EX-SC. The examples come from all three test datasets.
Fig~\ref{fig:GANs} demonstrates the ability of InfoPrint to detect the signatures of top-of-the-line
inpainting GANs. Unfortunately, no standard datasets exist to report quantitative results.
Furthermore, most of the examples are images that have been already processed, e.g.
resized or compressed (internet examples), which destroys camera-model traces, however, 
InfoPrint is still able to detect manipulations in these. This indicates that synthetically hallucinated pixels
carry a very different low-level signature which InfoPrint can detect and exploit.

\begin{figure}
	\centering
	\includegraphics[width=1\linewidth]{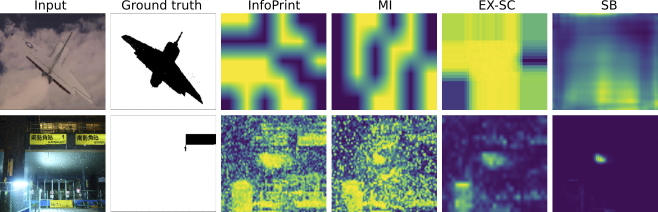}
	\caption{Failure examples. All methods tend to fail when the input image is small: e.g. $335\times251$ (row-1), or there are saturated regions (row-2). Log-probability maps of proposed methods are shown.}
	\label{fig:fail}
\end{figure}

However, there are always hard examples. In Fig~\ref{fig:fail}, we present two failure cases where 
all the compared algorithms fail to localize the correct tampered region. We identify two important causes
for this failure. In row-1, is a low-resolution image with dimensions $335\times251$ pixels. 
This means that the manipulated regions are also small, and the algorithms fail. Row-2 presents another typical
case, where the image contains strongly saturated regions. All the algorithms also fail in this case.

\section{Conclusion}

We presented a novel information theoretic formulation to address the issue of localizing tampered
regions (splices) in fake digital images. Using IB, we proposed an efficient variational solution 
and showed that it outperformed not only the standard expensive numerical method, but also published
forensic methods. Our IB formulation was also unique because we used it to learn noise-residual
patterns and suppress semantic contents rather than the other way around.
Additionally, we demonstrated our method's potential to detect inpainting operations by recent deep generative
methods. Finally, we identified failure cases such as saturated regions for future research directions.


\bibliographystyle{plain}

\begin{thebibliography}{10}
	
	\bibitem{achille:18}
	A.~Achille and S.~Soatto.
	\newblock Information dropout: Learning optimal representations through noisy
	computation.
	\newblock {\em IEEE Transactions on Pattern Analysis and Machine Intelligence},
	PP(99), 2018.
	
	\bibitem{agarwal:17}
	S.~{Agarwal} and H.~{Farid}.
	\newblock Photo forensics from {JPEG} dimples.
	\newblock In {\em 2017 IEEE Workshop on Information Forensics and Security
		{(WIFS)}}, pages 1--6, 12 2017.
	
	\bibitem{alemi:18}
	Alexander Alemi, Ben Poole, Ian Fischer, Joshua Dillon, Rif~A. Saurous, and
	Kevin Murphy.
	\newblock Fixing a broken {ELBO}.
	\newblock In {\em Proceedings of the 35th International Conference on Machine
		Learning {(ICML)}}, Jul 2018.
	
	\bibitem{alemi:17}
	Alexander~A. Alemi, Ian Fischer, Joshua~V. Dillon, and Kevin Murphy.
	\newblock Deep variational information bottleneck.
	\newblock In {\em International Conference on Learning Representations
		{(ICLR)}}, Apr 2017.
	
	\bibitem{ghosh:19}
	Anonymous.
	\newblock {SpliceRadar}: A learned method for blind image forensics.
	\newblock In {\em Supplied as supplementary material}, 2019.
	
	\bibitem{barber:03}
	David Barber and Felix~V. Agakov.
	\newblock The {IM} algorithm: A variational approach to information
	maximization.
	\newblock In {\em NeurIPS}, 2003.
	
	\bibitem{barni:17}
	Mauro Barni, Ehsan Nowroozi, and Benedetta Tondi.
	\newblock Higher-order, adversary-aware, double {JPEG}-detection via selected
	training on attacked samples.
	\newblock In {\em 25th European Signal Processing Conference ({EUSIPCO})},
	pages 281 -- 285, 08 2017.
	
	\bibitem{belghazi:18}
	Mohamed~Ishmael Belghazi, Aristide Baratin, Sai Rajeshwar, Sherjil Ozair,
	Yoshua Bengio, Aaron Courville, and Devon Hjelm.
	\newblock Mutual information neural estimation.
	\newblock In {\em Proceedings of the 35th International Conference on Machine
		Learning, {(ICML)}}, pages 531--540, 10--15 Jul 2018.
	
	\bibitem{bondi:17b}
	Luca Bondi, Silvia Lameri, David G\"{u}era, Paolo Bestagini, Edward Delp, and
	Stefano Tubaro.
	\newblock Tampering detection and localization through clustering of
	camera-based {CNN} features.
	\newblock In {\em The IEEE Conference on Computer Vision and Pattern
		Recognition (CVPR)}, pages 1855--1864, 07 2017.
	
	\bibitem{burgess:18}
	Christopher~P. {Burgess}, Irina {Higgins}, Arka {Pal}, Loic {Matthey}, Nick
	{Watters}, Guillaume {Desjardins}, and Alexander {Lerchner}.
	\newblock {Understanding disentangling in $\beta$-{VAE}}.
	\newblock {\em arXiv e-prints}, page arXiv:1804.03599, Apr 2018.
	
	\bibitem{chen:08}
	Mo~Chen, Jessica Fridrich, Miroslav Goljan, and Jan Lukás.
	\newblock Determining image origin and integrity using sensor noise.
	\newblock {\em Information Forensics and Security, IEEE Transactions on}, 3:74
	-- 90, 04 2008.
	
	\bibitem{chen:16}
	Xi~Chen, Yan Duan, Rein Houthooft, John Schulman, Ilya Sutskever, and Pieter
	Abbeel.
	\newblock {InfoGAN}: Interpretable representation learning by information
	maximizing generative adversarial nets.
	\newblock In {\em NeurIPS}, 2016.
	
	\bibitem{cozzolino:15}
	D.~{Cozzolino}, G.~{Poggi}, and L.~{Verdoliva}.
	\newblock Splicebuster: A new blind image splicing detector.
	\newblock In {\em 2015 IEEE International Workshop on Information Forensics and
		Security ({WIFS})}, pages 1--6, 11 2015.
	
	\bibitem{cozzolino:18b}
	Davide Cozzolino and Luisa Verdoliva.
	\newblock {Noiseprint: a CNN-based camera model fingerprint}.
	\newblock {\em arXiv}, 2018.
	
	\bibitem{carvalho:13}
	T.~J. d.~{Carvalho}, C.~{Riess}, E.~{Angelopoulou}, H.~{Pedrini}, and
	A.~d.~R.~{Rocha}.
	\newblock Exposing digital image forgeries by illumination color
	classification.
	\newblock {\em IEEE Transactions on Information Forensics and Security},
	8(7):1182--1194, 07 2013.
	
	\bibitem{hjelm:19}
	R~{Devon Hjelm}, Alex {Fedorov}, Samuel {Lavoie-Marchildon}, Karan {Grewal},
	Phil {Bachman}, Adam {Trischler}, and Yoshua {Bengio}.
	\newblock {Learning deep representations by mutual information estimation and
		maximization}.
	\newblock {\em arXiv e-prints}, page arXiv:1808.06670, Aug 2018.
	
	\bibitem{dirik:09}
	Ahmet~Emir Dirik and Nasir Memon.
	\newblock Image tamper detection based on demosaicing artifacts.
	\newblock In {\em in Proceedings of the 2009 IEEE International Conference on
		Image Processing (ICIP)}, 2009.
	
	\bibitem{medifor:17}
	Jonathan Fiscus, Haiying Guan, Yooyoung Lee, Amy Yates, Andrew Delgado, Daniel
	Zhou, David Joy, and August Pereira.
	\newblock {The 2017 Nimble Challenge Evaluation: Results and Future
		Directions}, 2017.
	
	\bibitem{fridrich:12}
	J.~{Fridrich} and J.~{Kodovsky}.
	\newblock Rich models for steganalysis of digital images.
	\newblock {\em IEEE Transactions on Information Forensics and Security},
	7(3):868--882, 06 2012.
	
	\bibitem{gloe:10}
	Thomas Gloe and Rainer Böhme.
	\newblock The `{D}resden {I}mage {D}atabase' for benchmarking digital image
	forensics.
	\newblock In {\em Proceedings of the 25th Symposium On Applied Computing (ACM
		SAC)}, volume~2, pages 1585--1591, 2010.
	
	\bibitem{goodfellow:2014}
	Ian~J. Goodfellow, Jean Pouget-Abadie, Mehdi Mirza, Bing Xu, David
	Warde-Farley, Sherjil Ozair, Aaron Courville, and Yoshua Bengio.
	\newblock Generative adversarial nets.
	\newblock In {\em NeurIPS}, 2014.
	
	\bibitem{he:16}
	K.~{He}, X.~{Zhang}, S.~{Ren}, and J.~{Sun}.
	\newblock Deep residual learning for image recognition.
	\newblock In {\em The IEEE Conference on Computer Vision and Pattern
		Recognition (CVPR)}, pages 770--778, 06 2016.
	
	\bibitem{higgins:17}
	Irina Higgins, Loic Matthey, Arka Pal, Christopher Burgess, Xavier Glorot,
	Matthew Botvinick, Shakir Mohamed, and Alexander Lerchner.
	\newblock $\beta$-{VAE}: Learning basic visual concepts with a constrained
	variational framework.
	\newblock In {\em International Conference on Learning Representations
		{(ICLR)}}, Apr 2017.
	
	\bibitem{huh:18}
	Minyoung Huh, Andrew Liu, Andrew Owens, and Alexei~A. Efros.
	\newblock Fighting fake news: Image splice detection via learned
	self-consistency.
	\newblock In Vittorio Ferrari, Martial Hebert, Cristian Sminchisescu, and Yair
	Weiss, editors, {\em Proceedings of the European Conference on Computer
		Vision {(ECCV)}}, pages 106--124, Cham, 2018. Springer International
	Publishing.
	
	\bibitem{kingma:13}
	Diederik~P. Kingma and Max Welling.
	\newblock Auto-encoding variational bayes.
	\newblock In {\em 2nd International Conference on Learning Representations,
		{(ICLR)} 2014, Banff, AB, Canada, April 14-16, 2014, Conference Track
		Proceedings}, 2014.
	
	\bibitem{liu:18}
	Guilin Liu, Fitsum~A Reda, Kevin~J Shih, Ting-Chun Wang, Andrew Tao, and Bryan
	Catanzaro.
	\newblock Image inpainting for irregular holes using partial convolutions.
	\newblock In {\em Proceedings of the European Conference on Computer Vision
		{(ECCV)}}, pages 85--100, 2018.
	
	\bibitem{lukas:06b}
	J.~Lukas, J.~Fridrich, and M.~Goljan.
	\newblock Detecting digital image forgeries using sensor pattern noise - art.
	no. 60720y.
	\newblock {\em Proceedings of SPIE - The International Society for Optical
		Engineering}, 6072:362--372, 02 2006.
	
	\bibitem{mahdian:09}
	Babak Mahdian and Stanislav Saic.
	\newblock Using noise inconsistencies for blind image forensics.
	\newblock {\em Image and Vision Computing}, 27(10):1497 -- 1503, 2009.
	\newblock Special Section: Computer Vision Methods for Ambient Intelligence.
	
	\bibitem{mayer:18}
	Owen Mayer and Mathew~C. Stamm.
	\newblock Learned forensic source similarity for unknown camera models.
	\newblock In {\em {IEEE International Conference on Acoustics, Speech and
			Signal Processing (ICASSP)}}. IEEE SigPort, 2018.
	
	\bibitem{nazeri:19}
	Kamyar Nazeri, Eric Ng, Tony Joseph, Faisal Qureshi, and Mehran Ebrahimi.
	\newblock Edgeconnect: Generative image inpainting with adversarial edge
	learning.
	\newblock {\em arXiv preprint arXiv:1901.00212}, 2019.
	
	\bibitem{otsu:79}
	N.~{Otsu}.
	\newblock A threshold selection method from gray-level histograms.
	\newblock {\em IEEE Transactions on Systems, Man, and Cybernetics},
	9(1):62--66, Jan 1979.
	
	\bibitem{peng:19}
	Xue~Bin {Peng}, Angjoo {Kanazawa}, Sam {Toyer}, Pieter {Abbeel}, and Sergey
	{Levine}.
	\newblock {Variational Discriminator Bottleneck: Improving Imitation Learning,
		Inverse RL, and GANs by Constraining Information Flow}.
	\newblock In {\em International Conference on Learning Representations
		{(ICLR)}}, May 2019.
	
	\bibitem{popescu:05}
	A.~C. {Popescu} and H.~{Farid}.
	\newblock Exposing digital forgeries in color filter array interpolated images.
	\newblock {\em IEEE Transactions on Signal Processing}, 53(10):3948--3959, 10
	2005.
	
	\bibitem{michael:18}
	Andrew~Michael Saxe, Yamini Bansal, Joel Dapello, Madhu Advani, Artemy
	Kolchinsky, Brendan~Daniel Tracey, and David~Daniel Cox.
	\newblock On the information bottleneck theory of deep learning.
	\newblock In {\em International Conference on Learning Representations
		{(ICLR)}}, May 2018.
	
	\bibitem{shwartz:17}
	Ravid {Shwartz-Ziv} and Naftali {Tishby}.
	\newblock {Opening the Black Box of Deep Neural Networks via Information}.
	\newblock {\em arXiv e-prints}, page arXiv:1703.00810, Mar 2017.
	
	\bibitem{tishby:99}
	Naftali Tishby, Fernando~C. Pereira, and William Bialek.
	\newblock The information bottleneck method.
	\newblock In {\em Proc. of the 37-th Annual Allerton Conference on
		Communication, Control and Computing}, pages 368--377, 1999.
	
	\bibitem{yu:18}
	Jiahui Yu, Zhe Lin, Jimei Yang, Xiaohui Shen, Xin Lu, and Thomas~S Huang.
	\newblock Generative image inpainting with contextual attention.
	\newblock In {\em Proceedings of the IEEE Conference on Computer Vision and
		Pattern Recognition {(CVPR)}}, pages 5505--5514, 2018.
	
	\bibitem{zampoglou:16}
	Markos Zampoglou, Symeon Papadopoulos, and Ioannis Kompatsiaris.
	\newblock Large-scale evaluation of splicing localization algorithms for web
	images.
	\newblock {\em Multimedia Tools and Applications}, 09 2016.
	
	\bibitem{zhou:18}
	Peng Zhou, Xintong Han, Vlad~I. Morariu, and Larry~S. Davis.
	\newblock Learning rich features for image manipulation detection.
	\newblock In {\em The IEEE Conference on Computer Vision and Pattern
		Recognition (CVPR)}, pages 1053--1061, 06 2018.
	
\end{thebibliography}

\end{document}